\documentclass[a4paper, 10pt, conference]{ieeeconf}      
\usepackage{FG2020}

\FGfinalcopy 

\IEEEoverridecommandlockouts                              
\usepackage{graphics} 
\usepackage{graphicx}
\usepackage{subcaption}
\usepackage{epsfig} 
\usepackage{mathptmx} 
\usepackage{times} 
\usepackage{amsmath} 
\usepackage{amssymb}  
\usepackage{diagbox}  
\usepackage{multirow} 
\usepackage{booktabs} 
\usepackage{xcolor}
\usepackage{algorithm}
\usepackage[noend]{algpseudocode}
\usepackage{multirow}
\usepackage{makecell}
\usepackage{helvet} 
\usepackage{courier}  
\usepackage[hyphens]{url}  
\usepackage{cite}

\def\FGPaperID{****} 

\title{\LARGE \bf
Emotion Recognition on large video dataset based on Convolutional Feature Extractor and Recurrent Neural Network
}

\author{\parbox{16cm}{\centering
    {\large Denis Rangulov, Muhammad Fahim}\\
    {\normalsize
    Innopolis University, Russia\\
    \{d.rangulov, m.fahim\}@innopolis.ru}}
}

\usepackage{hyperref}
\usepackage{siunitx}
\begin{document}

\ifFGfinal
\thispagestyle{empty}
\pagestyle{empty}
\else
\author{Anonymous FG2020 submission\\ Paper ID \FGPaperID \\}
\pagestyle{plain}
\fi
\maketitle

\begin{abstract}

For many years, the emotion recognition task has remained one of the most interesting and important problems in the field of human-computer interaction. In this study, we consider the emotion recognition task as a classification as well as a regression task by processing encoded emotions in different datasets using deep learning models. Our model combines convolutional neural network (CNN) with recurrent neural network (RNN) to predict dimensional emotions on video data. At the first step, CNN extracts feature vectors from video frames. In the second step, we fed these feature vectors to train RNN for exploiting the temporal dynamics of video. Furthermore, we analyzed how each neural network contributes to the system’s overall performance. The experiments are performed on publicly available datasets including the largest modern Aff-Wild2 database. It contains over sixty hours of video data. We discovered the problem of overfitting of the model on an unbalanced dataset with an illustrative example using confusion matrices. The problem is solved by downsampling technique to balance the dataset. By significantly decreasing training data, we balance the dataset, thereby, the overall performance of the model is improved. Hence, the study qualitatively describes the abilities of deep learning models exploring enough amount of data to predict facial emotions. Our proposed method is implemented using Tensorflow Keras. The code is publicly available in repository\footnote{\url{https://github.com/DenisRang/Combined-CNN-RNN-for-emotion-recognition}}.

\end{abstract}

\section{Introduction}

Emotions are the main indicators of human feelings. They can describe a distinct set of changes in physical state such as blood pressure rising or a particular face muscle moving. Several classical views of emotions suggest classification on a few basic categories like anger or happiness. Such views assume that each emotion has a defining underlying pattern in the brain and body. Another prospective assumes that the brain analyse past experience and predicts what the body should do in a similar situation. This theory is known as constructed emotions \cite{barrett2017emotions}. They are not triggered, on the contrary, we create them in our way. \par

Emotion capturing from videos is the simplest approach due to the simplicity of video recording in comparison with recording some physiological measurements such as EEG, blood pressure, and so on. Emotion recognition from videos has been a challenge for researcher community for many decades. Traditional way to encode emotions is developed by psychologist Paul Ekman \cite{eckman1972universal}. He divided all possible emotions into six basic categories: \textit{anger, disgust, fear, happiness, sadness, and surprise}. These emotions were selected because they were all perceived similarly regardless of culture. The prediction of these emotions already gives an abundant useful information about a person and enable further research of his or her condition. It is the oldest model of emotion recognition, so many existing technologies are based on this.\par

\begin{figure}[!ht]
 \centering
 \includegraphics[width=0.85\linewidth]{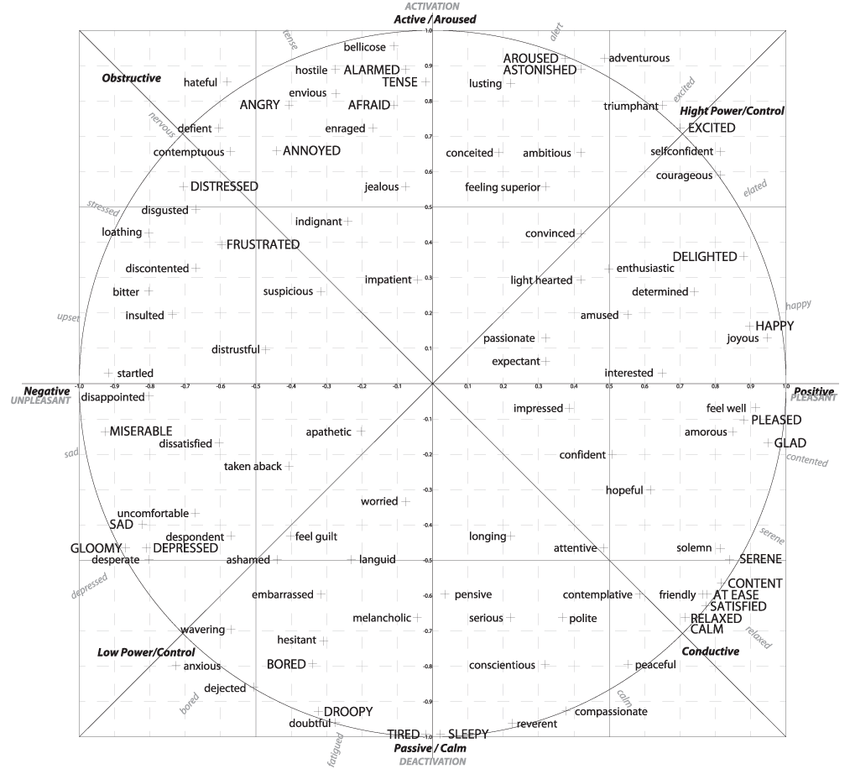}
 \caption{Dimensional structure for the semantic space of emotions \cite{scherer2005emotions}}
 \label{fig:2d-wheel}
\end{figure}

The main problem of previous encoding method is the constraint by only six categories. An effective way to deal with that can be to try to encode emotions in dimensional space to make it continuous. One coordinate shows how positive an emotion is and another tells how engaged or apathetic a subject appears. This way of encoding gives the ability to transform multiple dimensional coordinates to more complex emotion category. Such a complex emotion takes us closer to the theory of constructed emotions \cite{barrett2017emotions}. The second advantage is to output continuous emotion labels for video data containing continuous sequences of frames. Such model of emotion space is much more realistic and described in \cite{scherer2005emotions}.  Fig. \ref{fig:2d-wheel}, present the dimensional  emotions  in  valence-arousal space.
\par 

Prediction of emotions from videos is an analysis of sequences of frames for specific feature changes. These features can represent some facial muscles that can move in a particular order. Therefore, in every video particular emotional patterns are encoded. Features can be computed using traditional Computer Vision (CV) techniques \cite{o2019deep}. However, this way of solving emotion recognition task requires a significant effort from CV engineers such as deciding which features are important in each given frame for every kind of facial expression. As the number of emotions increases, feature extraction becomes more complicated.\par
On the other hand, the deep learning models can be fed with a whole dataset of annotated images. Thus, neural networks are able to discover the underlying patterns in video frames in a fast and accurate way. One of the main problems of such models for emotion recognition is the complexity of their architecture. If the complexity is too high, the model will take many computational resources for training and prediction. It is critical for embedded systems. The second consequence of using complex architectures is the possible overfitting of the model for recognizing emotions only from the subjects of the training set. In this research, we solve this challenge by creating a simple model with a low number of parameters to efficiently perform training and prediction of emotion recognition task.\par 

\section{Contribution and Paper Outline}
This study has the following contribution into research community:

\begin{itemize}
    \item A simple model to provide better results on large video dataset.
    
    \item Discover and address the problem of overfitting of the model on unbalanced dataset with an illustrative example using confusion matrices.
    
    \item Introduced data balancing technique by decreased number of training samples by 36\%, which contribute to improve the results .
    \item Finally, we provide analysis of neural networks by explaining how much the CNN and the RNN individually contribute to the overall performance.
\end{itemize}

The rest of this paper is organized as follows:  we briefly describe the related work in Section III.  Section IV provides the details about the proposed methodology.  In Section V, we presents experimental details followed by results and discussion in Section VI.  Finally,  Section VII provides conclusions and possible future directions.

\section{Related Work}

For dimensional emotions in valence-arousal space, the good way to feed input images to deep learning models is to convert them into sequences like usual order of frames in videos. Training on such data can exploit the temporal dynamics of the video. Feeding neural networks with the sequences can be done in different ways ad describes as follows:\par

\subsection{CNN and RNN trained jointly}

Kollias et al. \cite{kollias2018aff} developed an architecture by utilizing existing state-of-the-art CNN architectures together with different RNNs for Aff-Wild2 challenge. Nested CNN model are pretrained on different datasets with faces or random images. Then CNN and RNN are jointly trained in specific sequence to achieve good results on RECOLA\cite{recola} and Aff-Wild2\cite{kollias2019expression, kollias2018aff, kollias2018multi, kollias2019deep, zafeiriou2017aff, kollias2017recognition} datasets. This way of training has disadvantages such as the complexity of the overall combined model and therefore expensiveness of computational resources. Besides, they used attention mechanism by stacking an attention layer on top of the RNN. The attention mechanism deals with the problem of the limited short-term memory of RNNs. The aforementioned mechanism is a game-changing innovation that addressed this problem.\par

\subsection{CNN and RNN trained separately}

Another approach is presented in \cite{khorrami2017deep}, where author adapts CNN model to work with valence and arousal data from RECOLA dataset\cite{recola}. The last dense layer was changed with softmax activation for categorical emotions and regression layer to work with dimensional emotions. They proposed mechanism to incorporate the temporal information by using an RNN to propagate information from one time point to the next. Each input to the RNN is comprised of features from the second to the last fully connected layer of a single frame CNN. The main difference of this approach from previous is in separate training of CNN and RNN. They provide an evaluation of how much the CNN and the RNN individually contribute to the overall performance. They extend this approach to take audio and physiological data into consideration. He discovers usefulness of different features for prediction and examined how adding the audio and physiological features affected performance. Such types of data is not significant, if compared to video data.\par

Our approach utilized CNN architecture as a feature extractor and further connected with RNN gated recurrent units (GRU) to recognize the dimensional emotion in valence-arousal space. Furthermore, obtained results are better then the baseline model of Aff-Wild2 challenge.

\section{Methodology}
\begin{figure}[!hbtp]
 \centering
 \includegraphics[width=0.6\linewidth]{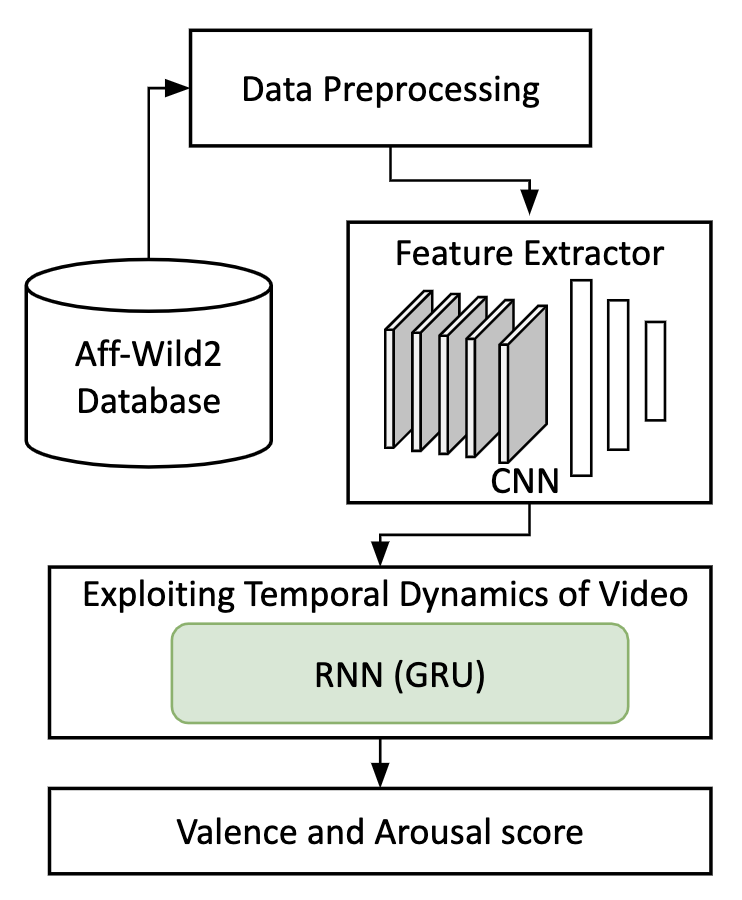}
 \caption{The block diagram of proposed model}
 \label{fig:model}
\end{figure}

The proposed system is presented in Fig. \ref{fig:model}. It consists of feature extractor and exploiting temporal dynamics of video unit to predict the valence and arousal score.

 The feature extractor is based on CNN architecture  \cite{khorrami2017deep} with few modifications. We choose this architecture because of its simplicity comparing to other complex ones, to achieve a trade-off between underfitting and overfitting. The CNN model is trained separately from RNN, to extract relevant feature vectors from images. Then RNN can learn useful temporal dynamic information of sequences of feature vectors. This approach takes lesser computation resources comparing with the others. The main reason is that we do not need to propagate frames every time during the training of the RNN. It allows us to use large batches with small feature vectors instead of large frames, which ensures better performance in terms of consumed memory.\par 

\subsection{Datasets}

In this research, we consider three different datasets to perform the experiments. The first Aff-Wild2 \cite{kollias2018aff} dataset is the largest dataset among other existing emotion datasets. According to train-validation partition and annotations provided by the ABAW 2020 challenge organizers \cite{kollias2020analysing}, there are 351 and 71 subjects in the training and validation subsets respectively for the valance-arousal estimation track. The number of samples with either low arousal or low valence is  small, because they are weakly expressed or negative emotions are rare as compared to highly expressed positive emotions. Hence the dataset is highly unbalanced.\par

The other two oldest and classical datasets the Extended Cohn-Kanade Dataset  (CK+)\cite{lucey2010extended} and Japanese Female Facial Expression (JAFFE) \cite{lyons_michael_1998_3451524} datasets are used for experiments to show how unbalancing in data can affect the overall performance of a deep learning model. Emotions are encoded categorically. This datasets are used in the most number of classical emotional recognition solutions. Subjects from two datasets have different ethnographies. CK+ consists of 593 sequences from 123 subjects. The validating  emotion  labels  were  only  assigned  to  327  sequences which  were  found  to  meet  criteria  for  one  of  seven  discrete  emotions. The Jaffe dataset are collected from 10 subjects. Each of them posed 3 or 4 examples of the six basic facial expressions (happiness, sadness, surprise, anger, disgust, fear) and a neutral face for a total of 219 images.

\subsection{Dataset preprocessing}

 We use all sequences from 118 subjects of CK+ dataset as in \cite{khorrami2017deep} experiments. For each sequence, the first image (i.e., neutral face) and three peak frames are used for prototypic expression recognition. For each image face is detected via Haar cascades \cite{jones2003fast} then cropped and resized to $96 \times 96$. We use gray-scaled images. The same preprocessing are used for JAFFE dataset. For Aff-Wild2 dataset, we are provided with all cropped-aligned face images of $112 \times 112$ dimensions, so we only need to resize it to $96 \times 96$. We exploit the full available data (training and validation). For training RNN, frames are combined into continuous sequences with length 100. Such window size is the best in \cite{khorrami2017deep}, during hyperparameter analysis experiments. Sequences are created only if there are no frames where a face has not been detected and both valence and arousal have been calculated. Thus, they do not have a gap between continuous frames. For frames where a model can not predict emotions, valence and arousal scores were later computed by linearly interpolating the scores from adjacent frames.\par
Besides, for Aff-Wild2 dataset, additional preprocessing is applied to reduce the overfitting. The overfitting is a possible issue that has bad influence on overall performance of our model for such an unbalanced dataset. We performed the following preprocessing to solve the challenge.\par

\subsubsection{Data augmentation}

We apply different data augmentations such as scaling, shifting, shearing, rotation, horizontal flips and changing brightness to make training data as diverse as possible. 

\subsubsection{Balancing dataset}

One way to balance a large dataset is to use downsampling. Firstly, we remove frames with both values of valence and arousal equal to zero. Secondly, we divide values of valence and arousal on 40 bins. Then we compute density of all 40x40 bins for all samples in the dataset. After that we select a frame for training with probability:

\begin{align}
\label{eq:select_frame_probability}
p=\frac{k_{current}}{k_{max}},
\end{align}
where $k_{current}$ is a number of samples in bin of this frame and $k_{max}$ is a number of samples in bin of the most frequent frame. $k_{max}$ was differently chosen for three settings:
\begin{enumerate}
    \item Number of samples in bin with maximum number of samples (subset 1).
    \item Mean number of samples among all bins (subset 2).
    \item Average between first two option (subset 3).
\end{enumerate}

\begin{figure}[h]
  \centering
  \begin{subfigure}[b]{0.49\linewidth}
    \includegraphics[width=\linewidth]{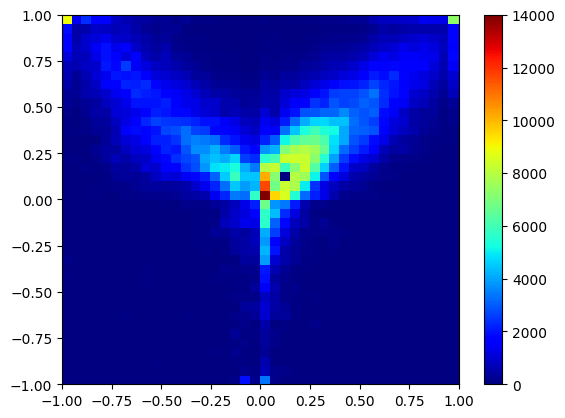}
     \caption{Subset 1 with \num{1267932} samples}
  \end{subfigure}
  \begin{subfigure}[b]{0.49\linewidth}
    \includegraphics[width=\linewidth]{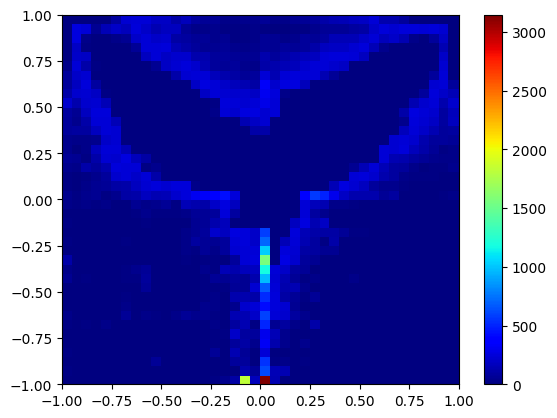}
    \caption{Subset 2 with \num{102360} samples}
  \end{subfigure}
  \begin{subfigure}[b]{0.49\linewidth}
    \includegraphics[width=\linewidth]{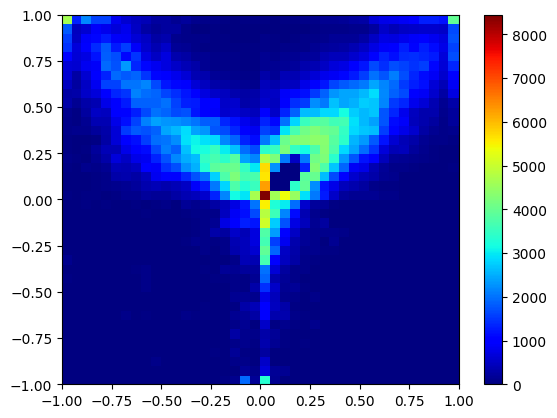}
    \caption{Subset 3 with \num{1030963} samples}
  \end{subfigure}
  \begin{subfigure}[b]{0.49\linewidth}
    \includegraphics[width=\linewidth]{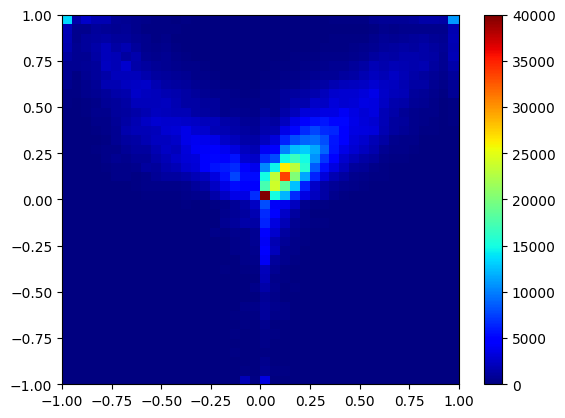}
    \caption{Whole dataset with \num{1620421} samples}
  \end{subfigure}
  \caption{2D valence-arousal histograms of Aff-Wild2 and downsampled subsets (horizontal axis is valance and vertical axis is arousal)}
  \label{fig:downsampling}
\end{figure}

As a result, we get different downsampled subsets of whole Aff-Wild2 dataset. They are shown in Fig. \ref{fig:downsampling}.\par

\begin{figure*}[h]
 \centering
 \includegraphics[width=\linewidth]{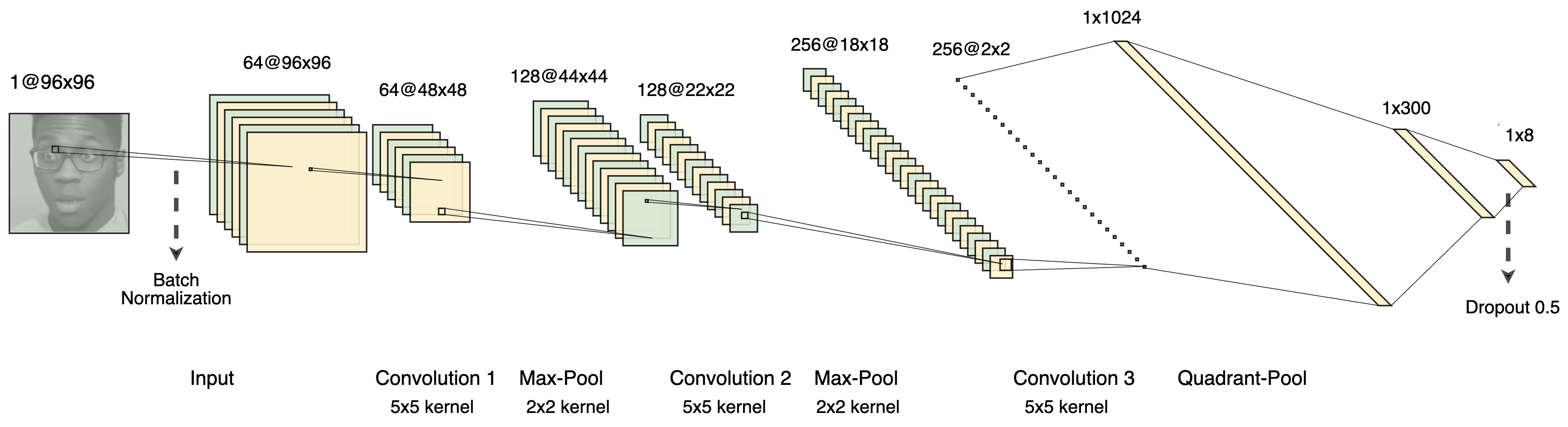}
 \caption{CNN architecture for emotion classification task on 8 categories}
 \label{fig:simplecnn}
\end{figure*}

\subsection{Feature extractor}

The main block of the overall system is feature extractor from images. Hence we evaluate different configuration of the CNN architecture proposed in \cite{khorrami2017deep} to solve a single frame regression task. Author solves the same emotion recognition task with smaller dataset. The best configuration is used further in RNNs training module. \par 

\subsubsection{CNN architecture}

Our CNN architecture is based on Khorrami \cite{khorrami2017deep}. He used a classical feed-forward convolutional neural network. In our networ, we added batch normalization first layer for zero-centering and normalizing each input. The network consist of three convolutional layers with 64, 128, and 256 filters. All filters have the same kernel size of $5 \times 5$ followed by Rectified Linear Unit(ReLU) activation functions. After the first two convolutional layers there are max pooling layers where quadrant pooling \cite{coates2011analysis} is applied after the third. Output of the quadrant pooling has $2 \times 2$ dimension per filter. Then a fully-connected layer with 300 hidden units is applied. This fully-connected layer is followed by dropout layer with probability 0.5.\par

Two different last layers are used for different tasks. The softmax layer is used for emotion classification on small datasets. This layer contains eight outputs corresponding to the number of expressions present in the CK+ training set. Architecture of the such model is in Fig. \ref{fig:simplecnn}. For regression task with data from Aff-Wild2 dataset, the dense layer with two units for predicting valence and arousal are used.\par

\subsubsection{Loss function}
A loss function is used to optimize the parameter values in a neural network model. In our model, Eq. \eqref{eq1} measures valence and arousal Concordance Correlation Coefficient(CCC) \cite{lawrence1989concordance} value and Eq. \eqref{eq2} is main loss function of dimensional emotion model.

\begin{equation}
\rho_{ccc} = \dfrac{2S_{xy}}{[s^2_x +s^2_y +(\overline x-\overline y)^2]}\label{eq1}
\end{equation}
\begin{equation}
\mathcal{L}_{ccc} = 1 - \dfrac{1}{2}[\rho_{a}+\rho_{v}]\label{eq2}
\end{equation}

\subsection{Exploiting of temporal dynamics of video data}

\begin{figure}[h]
 \centering
 \includegraphics[width=0.9\linewidth]{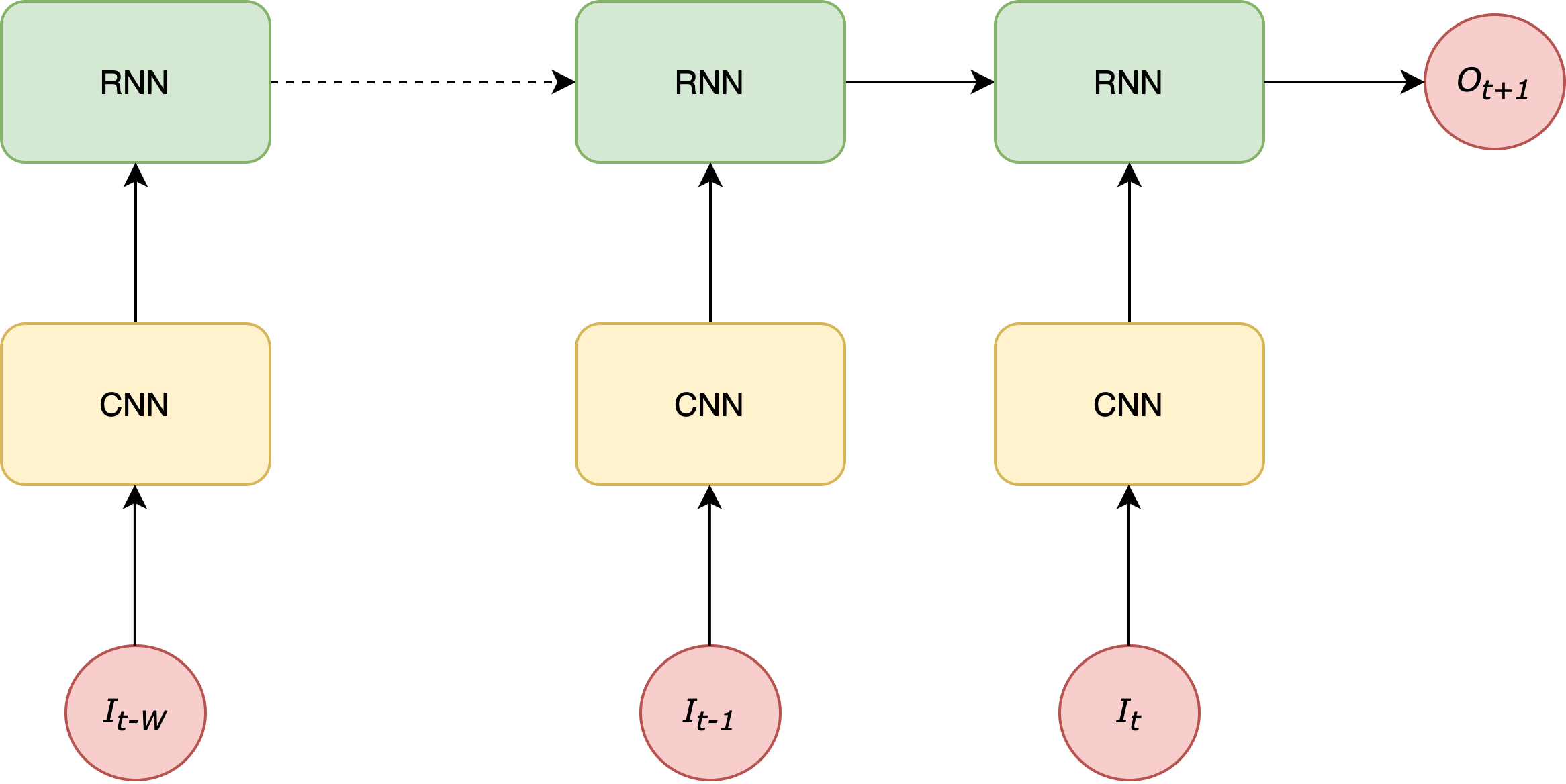}
 \caption{A combined CNN with RNN. Given a time \textit{t} in a video, we extract a window of length \textit{W} frames ([\textit{t} - \textit{W}, \textit{t}]) to predict valence and arousal at time \textit{t+1}}
 \label{fig:cnn_rnn_architecture}
\end{figure}

Our feature extractor CNN is trained on single frames so it completely ignores dynamics of changing emotions in sequence of frames in a single video. To address this issue we transform our dataset from videos to sequences of feature vectors and then fit RNN on it. We visualize such model in Fig. \ref{fig:cnn_rnn_architecture}.\par

We utlized the best performing number of units of RNN model from \cite{khorrami2017deep}. They had three hidden RNN layers with 100 hidden units in the first and second recurrent layers and 50 in the third. We compare it with architecture with single hidden RNN layer with 100 hidden units. We do not know how many frames we should remember to predict emotion in the next frame. It can be a few previous frames or all previous frames in a video. Therefore, we consider different types of RNN cells: simple RNN and GRU, which can capture short-term and long-term patterns in sequences, respectively. Final RNN models are represented in two configurations. The first one contains 1 layer with 100 hidden units. The second one contains 3 layers with 100 hidden units in the first two recurrent layers and 50 in the third. 

\section{Experiments Detail}

We utilized Google Colab to work with CK+ and JAFFE dataset, while GPU server with NVIDIA Tesla V100 (16 GB) graphics processor were used to process the large Aff-Wild2 dataset. Firstly, we validated the CNN model to get the similar results on small CK+ dataset as acheived in paper \cite{khorrami2017deep}. Then we evaluated this model on additional scopes by predicting emotion labels on unseen JAFFE dataset. Secondly, we experiment with large Aff-Wild2 dataset to find best configuration of feature extractor model. We compared separate predicting valence or arousal with predicting valence and arousal simultaneously by changing the number of units in the last dense layer of the CNN model from 1 to 2. Besides, we experimented with two loss functions: mean squared error (MSE) and difference between 1 and average CCC for valence and arousal described early. Lastly, we trained different RNN models on features from the best feature extractor model.\par
Two similar training strategies for the CNN model were used for emotion classification on CK+ dataset and regression tasks on Aff-Wild2 dataset. For all experiments we trained CNN and RNN models for 100 epochs with early stopping criteria. The batch size of $64$ and $128$ is used for CNN and RNN, respectively. They were trained from scratch using stochastic gradien descent (SGD) optimizer with momentum $0.9$, and weight decay parameter as $1e-5$. Along with that, we used constant learning rate of $0.01$. The parameters of each layer were initialized by default Xavier initialization or Glorot initialization strategy\cite{glorot2010understanding}. 

\section{Results and Discussion}

\subsection{Experiments on small CK+ and JAFFE datasets}

The CNN model trained over CK+ dataset was tested on test folder of CK+ and whole JAFFE dataset. The obtained results are presented in Table \ref{tab:ck_jaffe_accuracy}. As we can see, the accuracy on the CK+ dataset is much higher, that the one from the JAFFE dataset.\par

\begin{table}[h]
\centering
\caption{Results of experiments}
\label{tab:ck_jaffe_accuracy}
\begin{tabular}{|c||c|}
\hline
Dataset & Accuracy \\ \hline
\textbf{CK+}     & \textbf{95\%}     \\ \hline
JAFFE   & 50\%     \\ \hline
\end{tabular}
\end{table}

\begin{table}[h]
\centering
\caption{Number of samples with respect to emotional labels}
\label{tab:number_of_samples}
\resizebox{\columnwidth}{!}{%
\begin{tabular}{|c||c|c|c|c|c|c|c|c|}
\hline
                    & Neutral & Anger & Contempt & Disgust & Fear & Happy & Sadness & Surprise \\ \hline
Train folder of CK+ & \textbf{295}     & 129   & 51       & 153     & 60   & 186   & 84      & \textbf{219}      \\ \hline
Whole JAFFE         & 30      & 30    & -        & 29      & 32   & 31    & 31      & 30       \\ \hline
\end{tabular}%
}
\end{table}

Our thoughts were, as CK+ dataset was unbalanced and contains more examples of the Neutral, Surprise classes than the other classes. The number of training samples from CK+ and the number of all samples from JAFFE with respect to emotion labels are shown in Table \ref{tab:number_of_samples}.\par
\begin{figure}[H]
 \centering
 \includegraphics[width=\linewidth]{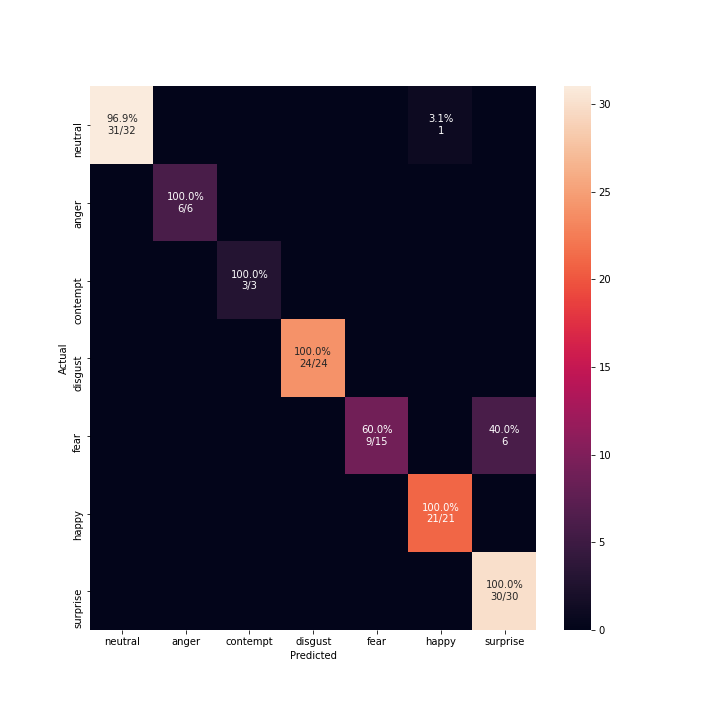}
 \caption{Confusion matrix over CK+ dataset}
 \label{fig:confusion_matrix_ck}
\end{figure}
\begin{figure}[H]
 \centering
 \includegraphics[width=\linewidth]{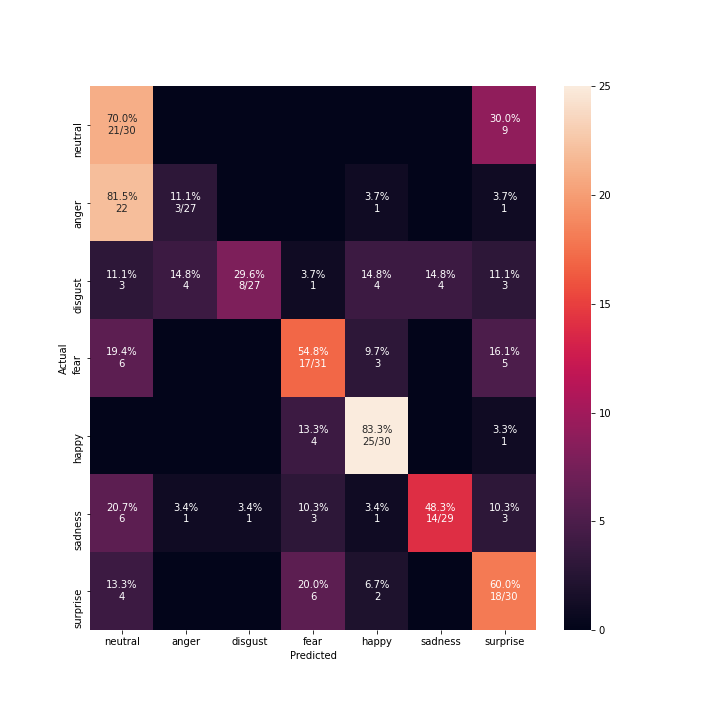}
 \caption{Confusion matrix over JAFFE dataset}
 \label{fig:confusion_matrix_jaffe}
\end{figure}

Thus, we also plotted the confusion matrix, to see, how well the model handles class with smaller amount of examples. In the Fig. \ref{fig:confusion_matrix_ck}, it is obvious that almost all classes, the model handles well for CK+. However, if we look on the confusion matrix of JAFFE dataset in the Fig. \ref{fig:confusion_matrix_jaffe}, we can see, that the imbalance of classes in the training resulted in the model finding more Neutral and Surprise classes, than others. It is easy to note how number of training samples correlates with errors in confusion matrix.\par

\subsection{Experiments on large Aff-Wild2 dataset}
      
The best results was obtained by the CNN predicting valence and arousal simultaneously with average CCC loss function. It is a little bit better than the baseline. We obtained a similar ratio between the CNN and the baseline model results with the ratio in \cite{khorrami2017deep} where the model was tested on the RECOLA dataset. Results are shown in Table \ref{tab:performance_comparison}. As we can see, it is sufficient to train model over subset 3 instead of the whole dataset. Hence, we achieved slightly better training results using only subset 3 with \num{1030963} samples versus all \num{1620421} samples. 

\begin{table}[h]
\centering
\caption{Performance comparison on Aff-Wild2 validation set}
\label{tab:performance_comparison}
\resizebox{\columnwidth}{!}{%
\begin{tabular}{|c||c|c|c|c|c|}
\hline
Method                      & \multicolumn{2}{c|}{RMSE}     & \multicolumn{2}{c|}{CCC}      & Best \# epochs \\ \hline
                            & Valence       & Arousal       & Valence       & Arousal       &                \\ \hline
Baseline                    & -             & -             & 0.14          & 0.24          & -              \\ \hline

CNN over whole Aff-Wild2    & 0.47          & 0.3           & 0.2           & 0.23          & 2              \\ \hline
CNN over subset 3           & 0.43          & 0.3           & 0.22          & 0.22          & 1              \\ \hline

CNN + Simple RNN, 1 layer   & 0.49          & 0.37          & 0.21          & 0.22          & 4              \\ \hline
CNN + GRU, 1 layer          & 0.51          & 0.32          & 0.2           & 0.35          & 8              \\ \hline
\textbf{CNN + GRU, 3 layer} & \textbf{0.45} & \textbf{0.29} & \textbf{0.23} & \textbf{0.39} & \textbf{17}    \\ \hline
\end{tabular}%
}
\end{table}

We selected the CNN trained on subset 3 for feature extraction. After that we trained different RNN on the extracted features from all images. There are \num{1496928} sequences with 100 feature vectors of size 300 in total. From the results, different RNNs can efficiently exploit the temporal dynamics of the data. Simple RNN shows worse results than GRU. Therefore, capturing long-term patterns is more efficient than capturing short-term patterns in small number of sequential frames in video sequence.

\section{Conclusion}

In this paper, we developed a deep neural networks for the task of emotion recognition. This task was split into two separate subtasks: feature extraction and exploiting of temporal dynamics of video data. We performed experiment with simple architecture on the largest modern video database Aff-Wild2. We discovered the problem of overfitting of the model on the unbalanced dataset with an illustrative example using confusion matrices. Next, we tried to balance the dataset by different downsampling. Therefore, we decreased the number of training samples on 36\% from \num{1620421} to \num{1030963} frames and achieved better performance of the model. We experimented with Aff-Wild2 database using the CNN and a combination of this model with different RNN models such as simple RNN and GRU. We obtain better results than baseline model with simple and general model, whose number of parameters is low. Therefore, such neural network can efficiently perform training and prediction emotion recognition task.\par
For future work, we are considering to adapt our model to multitask learning of different human affective behavior characteristics in-the-wild at the same time. These behavior characteristics are emotion encoded in one of seven basic categories, Action Units (AUs) describing all possible facial actions, and valence-arousal emotion encoding small changes in the intensity of each emotion on a continuous scale. 

\section{ACKNOWLEDGMENTS}

We gratefully acknowledge the support from D.Kollias with his discussion and providing cropped-aligned images.


\bibliography{FG-2020}
\bibliographystyle{ieee}

\end{document}